\documentclass[letterpaper, 10 pt, conference]{ieeeconf}  % Comment this line out if you need a4paper

\IEEEoverridecommandlockouts                              % This command is only needed if 
\title{\LARGE \bf
STDA-Net: Spectrogram-Based Domain Adaptation for cross-dataset Sleep Stage Classification
}

\author{Unaza Tallal$^{1}$, Shruti Kshirsagar$^{1}$, and Ankita Shukla$^{2}$% <-this % stops a space
\thanks{$^{1}$Unaza Tallal and Shruti Kshirsagar are with the School of Computing, Wichita State University, 1845 N. Fairmount, Wichita, KS 67260, USA
        {\tt\small uxtallal@shockers.wichita.edu, shruti.kshirsagar@wichita.edu}}%
\thanks{$^{2}$Ankita Shukla is with the Computer Science and Engineering Department, University of Nevada, Reno, NV, USA
        {\tt\small ankita.shukla@unr.edu}}%
}

\let\labelindent\relax
\usepackage{enumitem}
\usepackage{amsfonts}
\usepackage{multirow}
\usepackage{booktabs}
\usepackage{graphicx}
\usepackage{hyperref}
\usepackage{amsmath}
\usepackage{xcolor}

\begin{document}

\maketitle
\thispagestyle{empty}
\pagestyle{empty}

%%%%%%%%%%%%%%%%%%%%%%%%%%%%%%%%%%%%%%%%%%%%%%%%%%%%%%%%%%%%%%%%%%%%%%%%%%%%%%%%
\begin{abstract}

% Sleep stage classification across datasets remains challenging because EEG recordings vary in channel montages, sampling rates, recording environments, and subject populations. Most existing cross-dataset sleep-staging methods use one-dimensional EEG signals, whereas the use of two-dimensional spectrogram representations in unsupervised domain adaptation has remained unexplored. To address this gap, we propose STDA-Net (Spectrogram-based Temporal Domain Adaptation Network), a lightweight spectrogram-based temporal domain adaptation framework. The model combines a convolutional neural network (CNN) for spectrogram-based feature extraction, a bidirectional long short-term memory (BiLSTM) module for sequence modeling, and a domain-adversarial neural network (DANN) for source-target alignment. Experiments were conducted on Sleep-EDF, SHHS-1, and SHHS-2 under six cross-dataset transfer settings. Our proposed model achieved an average accuracy of 89.03\% and an average macro F1-score of 87.64\% across all six settings, showing consistently stronger balanced classification performance than the compared 1D baseline methods. In addition, the model produced lower variance across five independent runs, indicating improved stability and reproducibility. Overall, these findings indicate that a lightweight 2D spectrogram-based framework can effectively improve the robustness of cross-dataset sleep staging.
Accurate sleep stage classification across datasets remains challenging due to variability in EEG channel montages, sampling rates, recording environments, and subject populations. Although deep learning has shown considerable promise for automated sleep staging, most existing cross-dataset methods rely on one-dimensional EEG signal representations, whereas the use of two-dimensional spectrogram-based inputs within an unsupervised domain adaptation framework has remained largely unexplored. Here, we propose STDA-Net (Spectrogram-based Temporal Domain Adaptation Network), a framework that combines a convolutional neural network (CNN) for spectrogram-based feature extraction, a bidirectional long short-term memory (BiLSTM) module for temporal modeling of sleep dynamics, and a domain-adversarial neural network (DANN) for source-to-target feature alignment without requiring any labeled target-domain data during
training. Experiments are conducted on three publicly available datasets Sleep-EDF, SHHS-1, and SHHS-2  under six cross-dataset transfer settings. Results show that the proposed framework achieves an average accuracy of 89.03\% and an average macro F1-score of 87.64\%, consistently outperforming existing 1D baseline methods in terms of balanced classification performance, with substantially lower variance across five independent runs, indicating improved stability and reproducibility. Overall, these findings demonstrate that 2D spectrogram-based representations, combined with temporal modeling and adversarial domain adaptation, provide a robust and competitive alternative to conventional 1D EEG inputs for cross-dataset sleep staging.

Keywords— Sleep stage classification, EEG spectrograms, unsupervised
domain adaptation, cross-dataset generalization, adversarial learning, bidirectional LSTM, convolutional neural networks.

\end{abstract}

%%%%%%%%%%%%%%%%%%%%%%%%%%%%%%%%%%%%%%%%%%%%%%%%%%%%%%%%%%%%%%%%%%%%%%%%%%%%%%%%
\section{INTRODUCTION}
\label{sec:introduction}
Accurate sleep stage classification underlies the clinical diagnosis, monitoring, and management of sleep disorders, with broad implications for pharmaceutical evaluation, mental health monitoring, and the development of automated tools that reduce reliance on resource-intensive manual polysomnography scoring \cite{sankari2026sleepstudy, wara2025systematic}. While recent advances in deep learning have made significant improvements in automated staging systems, acquiring large annotated datasets for model training remains difficult in clinical settings \cite{heremans2022data}; this challenge becomes particularly critical in cross-dataset scenarios, where differences in subjects, sensors, and acquisition protocols cause domain shift and source-target mismatch \cite{he2023cross}. Although polysomnography (PSG) remains the gold standard for sleep monitoring, its multi-sensor setup can make data acquisition resource-intensive \cite{sankari2026sleepstudy}, motivating interest in more practical alternatives; electroencephalography (EEG) has, as a result, emerged as a widely used modality for sleep staging. It captures sleep-related brain activity using a reduced number of recording channels \cite{wara2025systematic}, with single-channel EEG in particular posing ongoing challenges for effective feature extraction and reliable automated classification \cite{zhou2022singlechannelnet}. Two-dimensional time-frequency representations such as spectrograms have been shown to provide complementary temporal and spectral information relevant to sleep staging \cite{tallal2026modulation}, and are explored here as the basis for feature modeling. However, a major limitation of existing deep learning approaches is their performance degradation under cross-dataset settings, where distributional differences across datasets lead to domain shift \cite{he2023cross}, and although domain adaptation, and transfer learning have emerged as effective strategies for improving model generalization \cite{ganin2016dann,heremans2022unsupervised}, limited attention has been given to domain adaptation for two-dimensional spectrogram-based cross-dataset sleep stage classification \cite{li2022deep}.

Motivated by this gap, this paper proposes STDA-Net (Spectrogram-based Temporal Domain Adaptation Network), a framework for unsupervised domain adaptation in sleep stage classification using 2D EEG spectrograms. The proposed framework integrates a CNN-based feature extractor, an unsupervised DANN module for domain-invariant feature learning \cite{ganin2016dann}, and a BiLSTM for temporal modeling of sleep dynamics, with the goal of improving cross-domain generalization while preserving the practical advantages of single-channel EEG-based sleep analysis. The key contributions of this work are as follows:

\begin{enumerate}
    \item We propose STDA-Net, a framework for cross-dataset sleep stage classification that operates on 2D EEG spectrograms. The architecture integrates three components: a CNN encoder that extracts per-epoch features from spectrograms, a BiLSTM that captures temporal dependencies across consecutive epochs, and adversarial domain adaptation (DANN) that aligns source and target feature distributions without requiring target labels.
    \item We conduct a systematic ablation study to quantify the individual contributions of temporal modeling, auxiliary supervision, and adversarial adaptation within the proposed framework
     \item Experiments conducted on three datasets demonstrate the robustness of the proposed framework specifically for a cross-dataset setting.
\end{enumerate}

The rest of this paper is organized as follows: Section \ref{relatedwork} reviews related work on domain adaptation for sleep analysis. Section \ref{sec:Proposed Method} presents the proposed methodology. Section \ref{sec: Experimental Setup} details the experimental setup, including the dataset description, and evaluation protocol. Section \ref{sec:RESULTS} presents results and discusses our findings, including ablation study results and baseline comparisons. Finally, Section \ref{sec:conclusion} provides the conclusion of the study. 

\section{Related Work}
\label{relatedwork}
Deep learning has become the dominant approach to automatic sleep-stage classification, with CNNs, transformers, and hybrid architectures demonstrating strong performance in EEG-based sleep analysis \cite{liu2024automatic}. Among existing approaches, single-channel EEG has received substantial attention due to its frequency-specific patterns useful for distinguishing sleep stages; for example, \cite{zhou2022singlechannelnet} proposed SingleChannelNet, an end-to-end model for sleep-stage classification using raw single-channel EEG without handcrafted features. Existing models are broadly grouped into 1D methods, trained directly on raw waveforms \cite{supratak2017deepsleepnet}, and 2D methods, which transform EEG into time-frequency representations such as spectrograms \cite{li2022deep}; comparative studies suggest that 2D representations can outperform certain 1D formulations, though their effectiveness depends on the dataset, signal representation, and model architecture \cite{li2022deep, haghayegh2023automated,kshirsagar2022quality,avila2019speech}.
Beyond label scarcity, model generalization across datasets has become a critical concern, as cross-dataset performance degradation arises from domain shift due to differences in subject populations, sensor variability, and channel configurations\cite{kshirsagar2022cross,  parupati2025towards,  kshirsagar2026geographic}. To address this, \cite{he2023cross} explored cross-scenario transfer learning with single-channel EEG and reported improved cross-domain performance, while \cite{wang2024generalizable} proposed SleepDG, a multi-level domain-alignment framework validated on five datasets. To reduce domain discrepancy, prior studies have predominantly relied on 1D signal representations; \cite{heremans2022unsupervised, mouradi2026robust} showed that adversarial domain adaptation effectively reduced the source-target gap in wearable sleep data and other domains as well, \cite{eldele2022adast} proposed ADAST, which combined an unshared attention mechanism with dual classifiers and iterative self-training to improve pseudo-label quality, and \cite{lyu2025deep} introduced DDAST, incorporating adaptive batch normalization and pseudo-label-based self-training for inter-subject alignment. Recently, EEG-based domain adaptation studies have improved cross-dataset sleep staging using adversarial joint alignment,  entropy-based learning, and active learning with minimal EEG channels \cite{ghasemigarjan2024optimizing, ghasemigarjan2025enhancing}. Nevertheless, the use of 2D spectrogram-based representations in unsupervised domain adaptation for sleep stage classification remains largely unexplored, motivating the proposed STDA-Net framework. 

\section{Proposed Methodology}
\label{sec:Proposed Method}

\begin{figure}[t]
    \centering
    \includegraphics[width=\columnwidth]{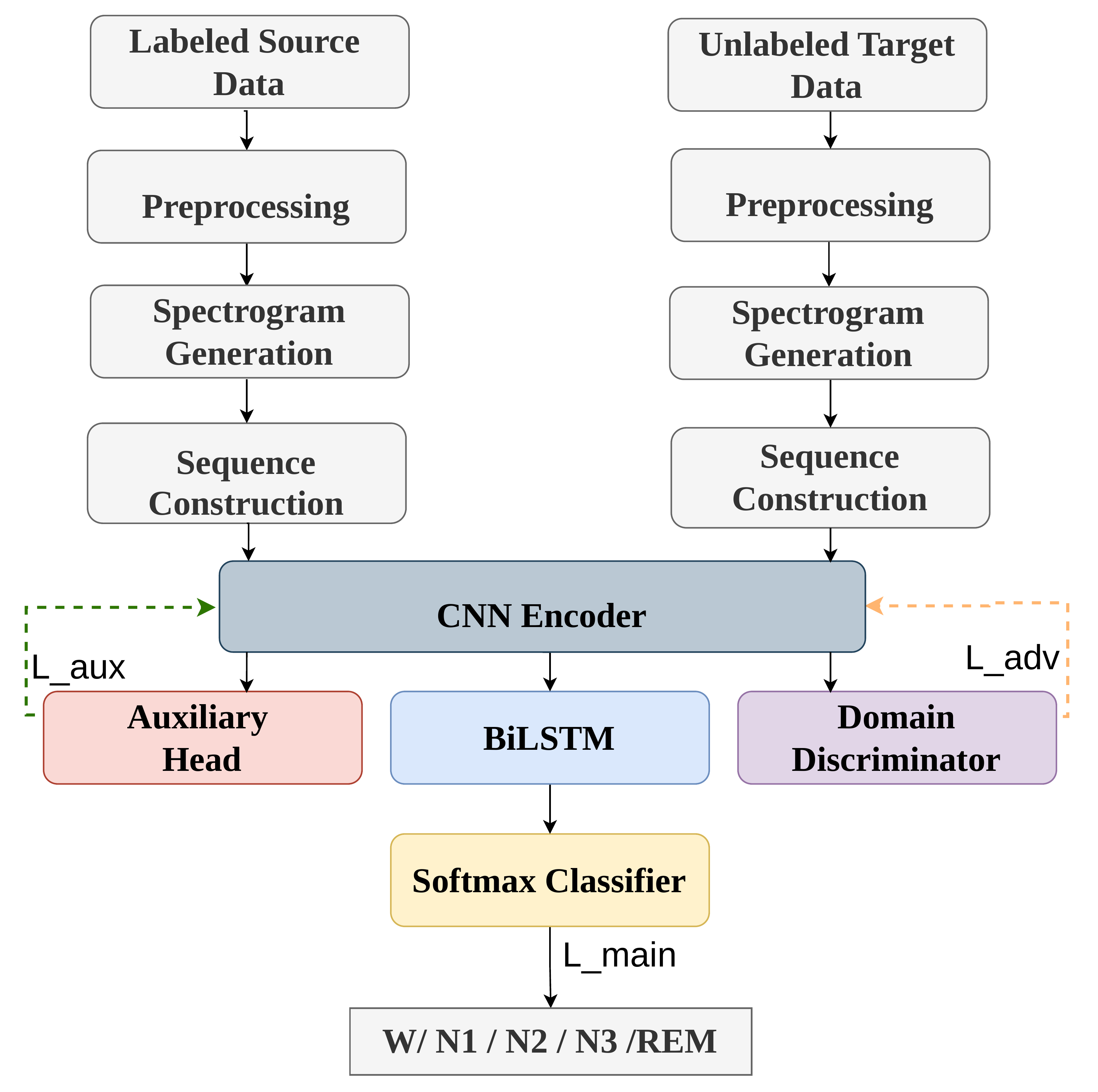}
    \caption{End-to-end pipeline of the proposed framework. Solid arrows indicate data flow. Dashed arrows indicate gradient flow during training.}
    \label{fig:Workflow}
\end{figure}

The proposed STDA-Net framework is illustrated in Fig.~\ref{fig:Workflow}. As can be seen, the framework consists of four interconnected components: a preprocessing and spectrogram generation module, a shared CNN epoch encoder, a BiLSTM temporal sequence module, and a domain-adversarial discriminator that uses an unsupervised domain adaptation setting, where labeled source-domain data and unlabeled target-domain data are used during training. The remainder of this section describes each component in detail.

 % The complete pipeline is illustrated in Fig. \ref{fig:Workflow} 
% integrates EEG
% signal spectrogram with sliding window sequence construction for texp
% \begin{enumerate}
%     \item spectrogram generation from raw EEG signals,
%     \item sliding window-based sequence construction for temporal context,
%     \item a CNN-based epoch encoder with an auxiliary classifier, and
%     \item a BiLSTM sequence classifier with domain adversarial training. 

% \end{enumerate}

\subsection{EEG Preprocessing and Spectrogram Generation}
\label{subsec:preprocess}
 
Raw single-channel EEG signals from source and target domains are processed independently to generate standardized 2D spectrograms. Signals from the C4-A1 channel in SHHS-1 and SHHS-2 are downsampled to 100~Hz to match the Fpz-Cz channel sampling rate in Sleep-EDF, and all signals are bandpass filtered between 0.5 and 30~Hz using a 4th-order Butterworth filter to preserve sleep-related frequency information. Following AASM standards, recordings are segmented into non-overlapping 30-s epochs, with deep sleep stages S3 and S4 merged into a single N3 class, unknown epochs excluded, and wake segments occurring more than 30 minutes before and after the sleep period removed. Z-score normalization is then applied to mitigate amplitude differences.
 
Each 30-s epoch is converted into a 2D time-frequency representation using the Short-Time Fourier Transform (STFT) with a Hamming window and 50\% overlap~\cite{xu2024sleep}, yielding a normalized single-channel spectrogram $x_i \in \mathbb{R}^{1 \times F \times T}$ of size $76 \times 60$. To model temporal dependencies across consecutive epochs, a sliding window of $L = 10$ epochs (corresponding to 5 minutes of continuous sleep) is applied with a stride of $S = 5$ epochs; sequences are constructed strictly within individual recording boundaries to avoid temporal leakage.

\subsection{CNN Epoch Encoder}
\label{subsec:cnn}
 
Each spectrogram in the input sequence is processed independently through a shared CNN backbone to extract discriminative epoch-level feature representations. The encoder consists of four convolutional blocks, each comprising a $3\times3$ convolution, batch normalization, and GELU activation, with stride-2 downsampling applied in the second and third blocks and a residual unit following each block to stabilize training. Global average pooling and a fully connected layer with dropout ($p = 0.3$) then project the features into a 128-dimensional vector:
\begin{equation}
    z_i = E_\theta(x_i) \in \mathbb{R}^{128}.
    \label{eq:encoder}
\end{equation}
An auxiliary classification head, consisting of two fully connected layers with GELU activation and dropout, is applied to these epoch-level features to provide direct per-epoch supervision and promote stable feature learning during joint optimization~\cite{li2022deep}:
\begin{equation}
    \hat{y}^{\text{aux}} = \text{softmax}\!\left(W_2 \cdot \text{GELU}(W_1 \cdot z_i + b_1) + b_2\right).
    \label{eq:aux}
\end{equation}
The encoder contains approximately 1.6 million parameters in total.

\subsection{BiLSTM Temporal Module}
\label{subsec:bilstm}
 
The sequence of epoch-level features $\mathbf{Z} = \{z_1, \ldots, z_L\}$ is fed into a two-layer bidirectional LSTM with a hidden size of 128 in each direction, producing context-aware representations:
\begin{equation}
    \mathbf{H} = \text{BiLSTM}(\mathbf{Z}) \in \mathbb{R}^{L \times 256},
    \label{eq:bilstm}
\end{equation}
where $h_t \in \mathbb{R}^{256}$ denotes the concatenation of forward and backward hidden states at time step $t$. A linear transformation followed by softmax then generates the final sleep stage prediction \cite{goodfellow2016deep}:
\begin{equation}
    \hat{y}_t^{\text{main}} = \text{softmax}(W_c h_t + b_c).
    \label{eq:classifier}
\end{equation}
Temporal modeling is particularly beneficial in the cross-dataset setting, as sleep-stage transition patterns for example, N1 rarely occurs between N3 and REM, and direct transitions from Wake to REM are uncommon are biologically constrained and relatively consistent across domains, thus naturally complementing the adversarial alignment objective.
\subsection{Domain-Adversarial Training}
\label{subsec:dann}
 
To reduce the domain gap between source and target CNN features, DANN-based adversarial training~\cite{ganin2016dann} is applied at the epoch feature level. A domain discriminator $D_\psi$, implemented as an MLP discriminator with two hidden layers of 128 and 64 units, ReLU activations, dropout, and a single sigmoid output, and is trained to distinguish source from target features, while the CNN encoder $E_\theta$ is trained to confuse the discriminator through a gradient reversal layer (GRL) \cite{ganin2016domain}:
\begin{equation}
    d_i = D_\psi\!\left(\text{GRL}_\lambda(z_i)\right).
    \label{eq:dann}
\end{equation}
Domain adaptation is applied at the per-epoch CNN feature level rather than at the BiLSTM output level, since the domain gap primarily arises in spectral feature representations, whereas temporal transition patterns tend to be relatively consistent across domains. The adaptation weight $\lambda$ follows a progressive schedule:
\begin{equation}
    \lambda = \frac{2}{1 + \exp(-10p)} - 1,
    \label{eq:lambda}
\end{equation}
where $p = \text{epoch}/\text{total\_epochs}$ increases from 0 to 1 during training, allowing the encoder to first learn discriminative source-domain features before gradually increasing the alignment pressure toward the target domain. 
\begin{table}[t]
\centering
\caption{Summary of datasets used in this study \cite{ghasemigarjan2025enhancing}.}
\label{tab:datasets}
\renewcommand{\arraystretch}{1.1}
\setlength{\tabcolsep}{4pt}
\small
\begin{tabular}{lcccc}
\hline
Dataset & Subjects & Channel & Fs (Hz) & Epoch (s) \\
\hline
Sleep-EDF & 20 & Fpz-Cz & 100 & 30 \\
SHHS-1  & 42 & C4-A1  & 125 & 30 \\
SHHS-2  & 44 & C4-A1  & 250 & 30 \\
\hline
\end{tabular}
\end{table}

\subsection{Optimization Objective} 

The total training loss combines three complementary objectives:
\begin{equation}
\mathcal{L}_{\mathrm{total}} = \mathcal{L}_{\mathrm{main}} + \alpha \,\mathcal{L}_{\mathrm{aux}} + \lambda \,\mathcal{L}_{\mathrm{adv}}
\label{eq:total_loss}
\end{equation}

The main classification loss $\mathcal{L}_{\text{main}}$ is the weighted cross-entropy calculated on BiLSTM output prediction using the labels from source-domain. The loss updates both the BiLSTM and the CNN encoder during backpropogation, train them jointly for temporal-context-aware sleep classification.
The auxiliary classification loss $\mathcal{L}_{\text{aux}}$ is also formulated as a weighted cross-entropy loss \cite{goodfellow2016deep} and is applied to per epoch CNN features using source labels. It provides direct intermediate supervision to CNN encoder and encourages the extracted spectrogram features to be discriminative durning training. The adversarial domain adaptation loss  \(\mathcal{L}_{\mathrm{adv}}\) is the binary cross-entropy loss from a domain discriminator that
classifies each CNN feature either originated from the source or target
domain. A gradient reversal layer (GRL) \cite{ganin2016domain},
inverts the gradient sign before it reaches the CNN encoder,
causing the encoder to produce features that are
indistinguishable between domains, \(\alpha = 0.5\) is auxiliary loss weight, and \(\lambda\) is the progressive DANN schedule defined in  Eq. (6). To address the class imbalance in sleep data, class weights are computer from source training set, with N1 representing only about 3\% of the total epochs.

During training, each mini-batch contains sequences from both the source and target domains. Source-domain sequences are used to compute the main classificaton loss \(\mathcal{L}_{\mathrm{main}}\), the auxiliary loss \(\mathcal{L}_{\mathrm{aux}}\), and the adversarial loss \(\mathcal{L}_{\mathrm{adv}}\). alignments. Target-domain sequences are used only in  \(\mathcal{L}_{\mathrm{adv}}\), as target labels are unavailable. During inference, predictions are generated using only the CNN encoder and BiLSTM classifier, while the auxiliary classifier and domain discriminator are removed.

\section{EXPERIMENTAL SETUP}
\label{sec: Experimental Setup}
Here, we describe our experimental setup, including the dataset, and the evaluation protocol.
\subsection{Datasets Description}
To evaluate the effectiveness of the proposed methodology, we conducted the experiments on three distinct and challenging datasets: Sleep-EDF1 (EDF) \cite{goldberger2000physionet, kemp2000sleepedf}, SHHS-1, and SHHS-2 \cite{zhang2018nsrr, quan1997shhs}. These datasets were selected because they differ in sampling rates, EEG channels, and cohort characteristics, making them suitable for assessing the robustness of the proposed framework under cross-dataset variability. Table \ref{tab:datasets} summarizes the key properties of the selected datasets. Notably, EDF uses the Fpz-Cz channel at 100 Hz, whereas SHHS-1 and SHHS-2 use the C4-A1 channel at 125 Hz and 250 Hz, respectively. We consider multiple cross-dataset settings to evaluate the proposed method under realistic domain shift, specifically when EEG is recorded from different brain regions, such as frontal (Fpz-Cz) and central (C4-A1) channels. To reduce the influence of respiratory events, only subjects with an apnea-hypopnea index (AHI) below 5 were included, following the dataset selection protocol adopted in the baseline study \cite{ghasemigarjan2025enhancing}. To ensure a fair comparison, subjects were selected using the same stratified split protocol. Such variability provides a challenging setting for assessing the proposed framework's ability to learn robust spectrogram-based representations.
\begin{figure}[!t]
    \centering
    \includegraphics[width=\columnwidth, trim=25 20 25 20, clip]{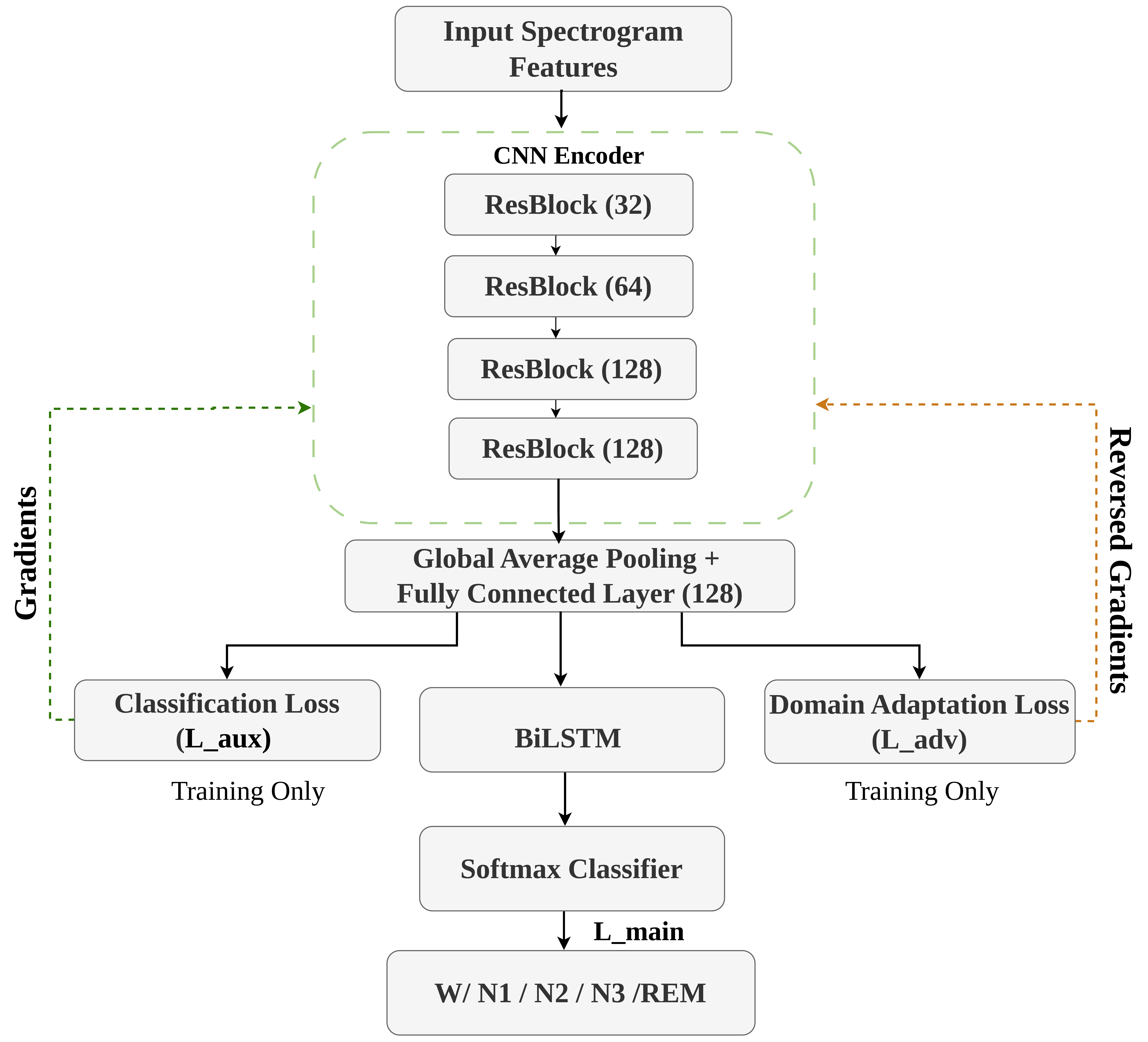}
    \caption{Overview of the proposed STDA-Net framework for unsupervised cross-dataset sleep stage classification.}
    \label{fig:architecture}
\end{figure}
A rigorous, leakage-free evaluation protocol is introduced here to ensure the validity and reproducibility of results, addressing three potential sources of data leakage. Subject-level leakage is prevented via subject-wise stratified splitting (60\% - 20\% - 20\% for training, validation, and testing); for Sleep-EDF subjects with two recording nights, both nights are assigned to the same split. Recording-boundary leakage is prevented by constructing sliding-window sequences strictly within individual recordings. Overlapping-window bias is addressed by averaging logits across all windows containing a given epoch, with the final label assigned via argmax, thus ensuring each physical epoch contributes only one prediction to the reported metrics. Target-domain labels remain unseen during training and are used only for evaluation. 
\begin{table}[!b]
\centering
\caption{Ablation study of the proposed framework. }
\label{tab:ablation}
\renewcommand{\arraystretch}{1.00}
\setlength{\tabcolsep}{3pt}
\scriptsize
\begin{tabular}{p{0.40\columnwidth}ccc}
\hline
\textbf{Components} & \textbf{Acc} & \textbf{MF1} & \textbf{$\kappa$} \\
\hline
CNN & 41.11 $\pm$ 3.76 & 34.25 $\pm$ 2.16 & 22.89 $\pm$ 2.27 \\
CNN + AUX & 45.69 $\pm$ 4.53 & 38.33 $\pm$ 4.41 & 27.85 $\pm$ 4.55 \\
CNN + DANN & 54.42 $\pm$ 1.48 & 48.03 $\pm$ 0.97 & 38.61 $\pm$ 1.68 \\
CNN + AUX + DANN & 53.56 $\pm$ 3.87 & 45.91 $\pm$ 2.44 & 37.31 $\pm$ 4.06 \\
CNN + AUX + BiLSTM & 79.30 $\pm$ 1.40 & 77.55 $\pm$ 2.08 & 68.33 $\pm$ 2.23 \\
\textbf{CNN + AUX + DANN + BiLSTM} & \textbf {87.69 $\pm$ 0.73} & \textbf {87.24 $\pm$ 0.74} & \textbf {81.86 $\pm$ 1.12} \\
\hline
\end{tabular}
\end{table}

\subsection{Evaluation Protocol}
Results are reported as mean~$\pm$~standard deviation across 5 independent runs with different random seeds (42, 52, 62, 72, 82), and model performance is evaluated using overall accuracy (Acc) that measures the proportion of correctly classified epochs, macro-averaged F1-score (MF1) to compute the unweighted average of the F1-scores across the five sleep stages, and Cohen's kappa coefficient~$\kappa$ to quantify inter-rater aggreement beyond chance, addressing class
imbalance in the label distribution \cite{phan2022automatic}. Early stopping is applied based on target validation MF1 with a patience of 10 epochs.

\section{RESULTS AND DISCUSSION}
\label{sec:RESULTS}

Here, we discuss our experimental results in light of existing literature.
\begin{table*}[!t]
\centering
\caption{Comparison of sleep stage classification \textbf{accuracy} (\%) under cross-dataset evaluation. Best results are shown in \textbf{bold}.}
\label{tab:comparison_acc}
\footnotesize
\setlength{\tabcolsep}{2.5pt}
\renewcommand{\arraystretch}{1.05}
\begin{tabular}{lccccccc}
\hline
\textbf{Method} & \textbf{EDF$\rightarrow$S1} & \textbf{EDF$\rightarrow$S2} & \textbf{S1$\rightarrow$EDF} & \textbf{S1$\rightarrow$S2} & \textbf{S2$\rightarrow$EDF} & \textbf{S2$\rightarrow$S1} & \textbf{Avg} \\
\hline

Deep CORAL \cite{sun2016deep}
& 69.53 $\pm$ 4.21 & 65.32 $\pm$ 3.26 & 77.46 $\pm$ 2.56 & 62.67 $\pm$ 1.82 & 69.73 $\pm$ 2.56 & 77.34 $\pm$ 2.45 & 69.67 \\
MDDA \cite{rahman2020minimum}       & 68.15 $\pm$ 0.65 & 70.43 $\pm$ 2.15 & 75.83 $\pm$ 3.21 & 68.82 $\pm$ 2.75 & 70.82 $\pm$ 1.44 & 77.45 $\pm$ 3.55 & 71.91 \\
DSAN \cite{zhu2020deep}       & 67.43 $\pm$ 0.72 & 75.64 $\pm$ 2.75 & 83.43 $\pm$ 0.53 & 69.23 $\pm$ 2.23 & 71.89 $\pm$ 2.35 & 75.80 $\pm$ 1.43 & 74.07 \\
DANN \cite{ganin2016dann}       & 68.72 $\pm$ 0.55 & 69.74 $\pm$ 2.53 & 78.32 $\pm$ 0.47 & 65.37 $\pm$ 0.27 & 68.85 $\pm$ 2.72 & 74.76 $\pm$ 0.37 & 70.96 \\
ADDA \cite{tzeng2017adversarial}       & 69.02 $\pm$ 2.36 & 65.72 $\pm$ 5.42 & 83.65 $\pm$ 2.61 & 58.76 $\pm$ 1.82 & 74.45 $\pm$ 1.38 & 74.53 $\pm$ 1.37 & 71.02 \\
CDAN \cite{long2018conditional}       & 66.19 $\pm$ 3.46 & 67.23 $\pm$ 1.45 & 79.12 $\pm$ 1.54 & 67.21 $\pm$ 3.18 & 74.56 $\pm$ 3.23 & 76.42 $\pm$ 1.88 & 71.78 \\
DIRT-T \cite{shu2018dirt}     & 68.73 $\pm$ 3.75 & 61.42 $\pm$ 3.14 & 80.78 $\pm$ 1.25 & 67.16 $\pm$ 2.75 & 74.75 $\pm$ 2.35 & 79.28 $\pm$ 1.72 & 72.02 \\
ADAST \cite{eldele2022adast}      & 77.53 $\pm$ 2.33 & 68.57 $\pm$ 2.36 & 77.84 $\pm$ 4.21 & 73.27 $\pm$ 1.06 & 78.48 $\pm$ 2.73 & 80.20 $\pm$ 1.34 & 75.98 \\
ADJLDA \cite{ghasemigarjan2024optimizing}     & 86.32 $\pm$ 1.35 & 85.89 $\pm$ 1.26 & 84.56 $\pm$ 2.21 & 85.47 $\pm$ 2.21 & 87.01 $\pm$ 1.45 & 88.38 $\pm$ 2.16 & 86.25 \\
\textbf{ADAADL \cite{ghasemigarjan2025enhancing}}& \textbf{88.32 $\pm$ 1.35}
& \textbf{87.89 $\pm$ 1.26}
& {86.56 $\pm$ 2.21} 
& \textbf{87.47 $\pm$ 2.21} 
& {89.01 $\pm$ 1.45} 
& {90.38 $\pm$ 2.16} 
& {88.25}  \\
\hline
\textbf{STDA-Net (Proposed)}
& 87.69 $\pm$ 0.73 
& 82.32 $\pm$ 1.37 
& \textbf{92.99 $\pm$ 1.13}
& 85.10 $\pm$ 0.74 
& \textbf{93.29 $\pm$ 0.97} 
& \textbf{92.79 $\pm$ 0.99} 
& \textbf{89.03} \\
\hline

\end{tabular}
\end{table*}

\begin{table*}[!t]
\centering
\caption{Comparison of sleep stage classification \textbf{Macro-F1} (\%) under cross-dataset evaluation. Best results are shown in \textbf{bold}.}
\label{tab:comparison_F1}
\footnotesize
\setlength{\tabcolsep}{2.5pt}
\renewcommand{\arraystretch}{1.05}
\begin{tabular}{lccccccc}
\hline
\textbf{Method} & \textbf{EDF$\rightarrow$S1} & \textbf{EDF$\rightarrow$S2} & \textbf{S1$\rightarrow$EDF} & \textbf{S1$\rightarrow$S2} & \textbf{S2$\rightarrow$EDF} & \textbf{S2$\rightarrow$S1} & \textbf{Avg} \\
\hline

Deep CORAL \cite{sun2016deep}  & 58.34 $\pm$ 3.58 & 45.65 $\pm$ 4.63 & 65.96 $\pm$ 2.39 & 53.44 $\pm$ 1.57 & 60.25 $\pm$ 1.31 & 65.34 $\pm$ 2.35 & 58.16 \\
MDDA  \cite{rahman2020minimum}      & 58.12 $\pm$ 0.47 & 52.17 $\pm$ 2.68 & 65.78 $\pm$ 2.56 & 57.34 $\pm$ 1.63 & 61.13 $\pm$ 1.21 & 65.13 $\pm$ 1.78 & 59.94 \\
DSAN   \cite{zhu2020deep}     & 58.42 $\pm$ 0.51 & 57.17 $\pm$ 3.21 & 69.48 $\pm$ 1.32 & 58.71 $\pm$ 2.47 & 61.29 $\pm$ 3.14 & 65.21 $\pm$ 1.56 & 61.71 \\
DANN    \cite{ganin2016dann}    & 57.89 $\pm$ 0.34 & 52.82 $\pm$ 3.43 & 66.67 $\pm$ 1.53 & 56.28 $\pm$ 1.28 & 63.34 $\pm$ 1.15 & 63.33 $\pm$ 1.56 & 60.01 \\
ADDA  \cite{tzeng2017adversarial}     & 60.16 $\pm$ 1.37 & 46.86 $\pm$ 5.46 & 71.21 $\pm$ 2.35 & 53.63 $\pm$ 1.57 & 61.73 $\pm$ 1.58 & 61.34 $\pm$ 1.38 & 59.15 \\
CDAN  \cite{long2018conditional}    & 55.73 $\pm$ 4.65 & 54.32 $\pm$ 1.47 & 66.43 $\pm$ 2.25 & 57.69 $\pm$ 2.38 & 61.71 $\pm$ 1.56 & 66.56 $\pm$ 2.65 & 60.40 \\
DIRT-T  \cite{shu2018dirt}    & 58.48 $\pm$ 4.73 & 50.22 $\pm$ 5.65 & 68.34 $\pm$ 1.58 & 58.39 $\pm$ 1.47 & 59.73 $\pm$ 3.33 & 65.35 $\pm$ 1.53 & 60.08 \\
ADAST   \cite{eldele2022adast}    & 63.82 $\pm$ 1.75 & 55.62 $\pm$ 3.24 & 66.21 $\pm$ 2.17 & 61.29 $\pm$ 1.32 & 67.51 $\pm$ 2.21 & 67.89 $\pm$ 1.26 & 63.72 \\
ADJLDA  \cite{ghasemigarjan2024optimizing}    & 77.32 $\pm$ 2.21 & 75.39 $\pm$ 4.22 & 77.16 $\pm$ 3.45 & 76.37 $\pm$ 1.78 & 79.65 $\pm$ 2.37 & 81.19 $\pm$ 1.74 & 78.25 \\
ADAADL \cite{ghasemigarjan2025enhancing}
& {80.32 $\pm$ 2.21}
& {78.39 $\pm$ 4.22}
& {80.16 $\pm$ 3.45}
& {79.37 $\pm$ 1.78}
&  {82.65 $\pm$ 2.37}
&  {84.19 $\pm$ 1.74}
&  {80.84} \\
\hline

\textbf{STDA-Net (Proposed)}
&  \textbf{87.24 $\pm$ 0.74}
&  \textbf{79.26 $\pm$ 1.68}
&  \textbf{91.59 $\pm$ 0.99}
&  \textbf{83.41 $\pm$ 0.84}
&  \textbf{91.78 $\pm$ 1.02}
&  \textbf{92.60 $\pm$ 0.76}
&  \textbf{87.64} \\
\hline
\end{tabular}
\end{table*}

%-------------------------------------------------------------
\subsection{Ablation Study}
A systematic ablation study is conducted here to evaluate the contribution of each architectural component, with six model variants compared in Table~\ref{tab:ablation}. All variants share the same cross-dataset setup (Source: Sleep-EDF, Target: SHHS-1), CNN backbone, and subject-wise splits to ensure a fair comparison.
 It can be observed clearly that adding an auxiliary head to the CNN backbone improves performance, suggesting that direct epoch-level supervision helps the encoder learn discriminative spectral features, consistent with findings in~\cite{li2022deep}. Similarly, the CNN~+~DANN configuration outperforms the CNN-only baseline, confirming a clear domain gap between source and target datasets and demonstrating that adversarial alignment helps reduce it~\cite{ganin2016dann}. However, the CNN~+~AUX~+~DANN setting shows a slight performance decrease across all three metrics, suggesting that without temporal modeling, auxiliary supervision and adversarial alignment objectives may interfere with each other during epoch-level training. Adding a BiLSTM for temporal modeling yields the largest individual gain improving accuracy by 25.74\%, MF1 by 31.64\%, and $\kappa$ by 31.02\% over the CNN~+~AUX baseline highlighting the importance of sequential context for capturing sleep-stage transitions, consistent with the role of temporal modeling reported in prior sleep staging work~\cite{supratak2017deepsleepnet}. When DANN is further added, the full proposed model gains an additional 8.39\% in accuracy, 9.69\% in MF1, and 13.53\% in $\kappa$, demonstrating that the best cross-dataset performance is achieved by combining CNN features, auxiliary supervision, adversarial alignment, and temporal modeling.

\subsection{Comparison with Baseline Methods}
 
Tables~\ref{tab:comparison_acc} and~\ref{tab:comparison_F1} compare STDA-Net with state-of-the-art cross-dataset sleep staging methods. Unlike existing baselines that rely on 1D EEG signal representations, the proposed method performs classification using 2D EEG spectrograms. As can be seen, the proposed framework achieves consistently higher macro F1 across all six transfer settings than the 1D baselines, indicating more balanced classification across sleep stages rather than performance biased by majority classes, an important distinction in sleep staging, where dominant stages such as Wake and N2 can inflate overall accuracy while minority stages such as N1 remain challenging to classify~\cite{eldele2022adast}. The proposed method also demonstrates substantially lower variance across five independent runs than the 1D baselines, suggesting that 2D spectrogram representations reduce within-subject variability and that the observed gains are real rather than inflated by a specific data split. It also indicates improved stability and reproducibility across cross-datasets. 
Several additional insights emerge from these results. The consistently higher macro F1 of STDA-Net, despite minor accuracy drops in some directions, indicates that 2D spectrogram representations provide a more balanced characterization of sleep stages than 1D baselines, which tend to favor dominant classes~\cite{he2023cross}. An interesting directional asymmetry is also observed: transfer from SHHS to Sleep-EDF consistently outperforms the reverse direction i.e., S1→EDF outperforms EDF→S1 by 5.3\%, and S2→EDF outperforms EDF→S2 by 10.97\%, suggesting that training on a physiologically diverse source population yields richer and more generalizable spectral representations~\cite{wang2024generalizable}, with practical implications for clinical deployment. 
\section{Conclusion}
\label{sec:conclusion}

In this paper, we investigated the effectiveness of 2D EEG spectrograms as an alternative to 1D representations for improving cross-dataset sleep staging. To this end, we proposed STDA-Net, an unsupervised domain adaptation framework that combines CNN-based spectrogram feature extraction, BiLSTM-based temporal modeling, and DANN-based adversarial alignment to reduce the discrepancy
between source and target EEG datasets. Experiments conducted across Sleep-EDF, SHHS-1, and SHHS-2 in six cross-dataset transfer settings showed that the proposed framework achieves consistently higher macro F1 than existing 1D baselines, with substantially lower variance
across five independent runs, indicating improved stability and reproducibility. These results demonstrate that 2D time-frequency representations, combined with temporal modeling and adversarial domain adaptation, provide a
robust and competitive framework for cross-dataset sleep staging. Future work will focus on validating STDA-Net on additional databases with greater population variability, such as ISRUC and MASS, and on extending the framework to multi-channel EEG configurations to incorporate spatial
information that complements the learned temporal and spectral features.
The implementation details of the proposed method are publicly available at:
\href{https://github.com/Unaiza4/STDA-Net-Spectrogram-Based-Temporal-Domain-Adaptation-for-Cross-Dataset-Sleep-Stage-Classification}{GitHub repository}.
\section*{Acknowledgment}
This work was supported by the NSF Institute for Foundations of Machine Learning (IFML) and used the NCSA Delta GPU through allocation CIS251429 from the ACCESS program
\cite{boerner2023access}.

\addtolength{\textheight}{-12cm}   % This command serves to balance the column lengths
                                  % on the last page of the document manually. It shortens
                                  % the textheight of the last page by a suitable amount.
                                  % This command does not take effect until the next page
                                  % so it should come on the page before the last. Make
                                  % sure that you do not shorten the textheight too much.

%%%%%%%%%%%%%%%%%%%%%%%%%%%%%%%%%%%%%%%%%%%%%%%%%%%%%%%%%%%%%%%%%%%%%%%%%%%%%%%%

%%%%%%%%%%%%%%%%%%%%%%%%%%%%%%%%%%%%%%%%%%%%%%%%%%%%%%%%%%%%%%%%%%%%%%%%%%%%%%%%

%%%%%%%%%%%%%%%%%%%%%%%%%%%%%%%%%%%%%%%%%%%%%%%%%%%%%%%%%%%%%%%%%%%%%%%%%%%%%%%%

%%%%%%%%%%%%%%%%%%%%%%%%%%%%%%%%%%%%%%%%%%%%%%%%%%%%%%%%%%%%%%%%%%%%%%%%%%%%%%%%

\bibliographystyle{IEEEtran}

\bibliography{references}

\end{document}